\documentclass{InsightArticle}
\usepackage{graphicx}

\usepackage{xcolor}
\definecolor{mygreen}{rgb}{0,0.6,0}
\definecolor{mygray}{rgb}{0.5,0.5,0.5}
\definecolor{mymauve}{rgb}{0.58,0,0.82}
\definecolor{light-gray}{gray}{0.95} 
\usepackage[draft]{minted} 

\setminted[c++]{%
  fontsize=\small,
  style=borland,
  breaklines,
  bgcolor=light-gray
}
\newcommand{\inline}[1]{\mintinline{c++}/#1/}

\newcommand{\github}[1]{\href{https://github.com/phcerdan/ITKIsotropicWavelets/blob/master/include/itk#1.h}{itk::#1}}
\usepackage{amssymb}
\usepackage{amsmath}
\usepackage{amsthm}
\usepackage{physics} 
\usepackage{enumitem}
\theoremstyle{definition}
\newtheorem{definition}{Definition}[section]
\newtheorem{proposition}{Proposition}

\newcommand{\conj}[1]{#1^{\ast}}
\newcommand{\fft}[1]{\hat{#1}}
\newcommand{\trans}[1]{#1^{\mathsf{T}}}
\def\w{\omega} 
\def\vn{\bm{n}} 
\def\vx{\bm{x}} 
\usepackage{mathtools} 
\usepackage{bm} 
\usepackage{float}
\usepackage{caption} 
\usepackage{subcaption}
\usepackage{pgf} 
\graphicspath{ {images/}{figures/} }
\usepackage{tikz}
\usepackage{csquotes}
\usepackage[
bookmarks,
bookmarksopen,
backref,
colorlinks,linkcolor={blue},citecolor={blue},urlcolor={blue},
]{hyperref}

\title{Isotropic and Steerable Wavelets in N Dimensions. A multiresolution analysis framework.}

\newcommand{\IJhandlerIDnumber}{3558}

\release{0.4}

\author{P. H. Cerdan $^{1,2}$}
\authoraddress{$^{1}$ Institute of Fundamental Sciences, Massey University, New Zealand\\
  $^{2}$ p.hernandezcerdan@massey.ac.nz}

\begin{document}


\maketitle

\ifhtml
\chapter*{Front Matter\label{front}}
\fi

\begin{abstract}
\noindent
This document describes the implementation of the external module ITKIsotropicWavelets, a multiresolution (MRA) analysis framework using isotropic and steerable wavelets in the frequency domain. This framework provides the backbone for state of the art filters for denoising, feature detection or phase analysis in N-dimensions. It focus on reusability, and highly decoupled modules for easy extension and implementation of new filters, and it contains a filter for multiresolution phase analysis,

The backbone of the multi-scale analysis is provided by an isotropic band-limited wavelet pyramid, and the detection of directional features is provided by coupling the pyramid with a generalized Riesz transform.
The generalized Riesz transform of order N behaves like a smoothed version of the Nth order derivatives of the signal. Also, it is steerable: its components impulse responses can be rotated to any spatial orientation, reducing computation time when detecting directional features.

This paper is accompanied with the source code, input data, parameters and
output data that the author used for validating the algorithm described in
this paper. This adheres to the fundamental principle that scientific
publications must facilitate reproducibility of the reported results.
\end{abstract}

\IJhandlenote{\IJhandlerIDnumber}

\tableofcontents

\section{Wavelet Multiresolution Analysis}
\label{sec:first}

\subsection{Introduction}
\label{sub:introduction}
This module is a generalization to N dimension for ITK of the work made by: \cite{freeman_design_1991, simoncelli_steerable_1995, kovesi_image_1999, held_steerable_2010,chenouard_3d_2012, unser_steerable_2011, pad_vow:_2014}.

Learning about wavelets from research papers is not easy. The topic is rich and deep, the same tool has spanned different research fields, from theoretical physics, to seismology and signal processing.
Yves Meyer has won the Gauss Price (2010), and Abel Price (2017) in Mathematics for ``his pivotal role in the development of wavelets and multi-resolution analysis''. The detection of Gravitational waves with the LIGO experiment involved the use of wavelets to analyse the signals.

Stephan Mallat \cite{mallat_theory_1989} had also a fundamental role to develop the Multiresolution Analysis (MRA) and make the implementation available for applications.

The steerable framework \cite{freeman_design_1991, simoncelli_steerable_1995} is widely used in applications to rotate the filter bank to any direction, avoiding expensive computations.

We will also implement more recent work \cite{held_steerable_2010, unser_steerable_2011, chenouard_3d_2012} that uses the Riesz transform (a natural generalization of the Hilbert transform for N dimensions) to provide a generalization and extra flexibility to the steerable pyramid.

\subsection{Motivation: spatial and frequency resolution}
\label{sub:Motivation}
Wavelets share the same motivation than the short-time FFT, or the windowed Fourier transform: get a representation of the signal/image/function that is well localized in space \textbf{and} frequency domains. Heisenberg Uncertainty principle applies here and it is called the Heisenberg-Gabor limit: $\Delta t \cdot \Delta f \geq \frac{1}{4\pi}$, limiting the simultaneous resolution of a signal in time-frequency.
The Fourier transform, which is the representation of the signal in the basis of harmonic functions $\{\sin(f),\cos(f)\}\ \forall f \in \mathbb{R}$, is completely localized in frequency $\Delta f = 0$, but has global support in space, i.e infinitely de-localized in space $\Delta t = \infinity$.

\begin{figure}[H]
  \centering
  \resizebox{0.9\textwidth}{!}{\input{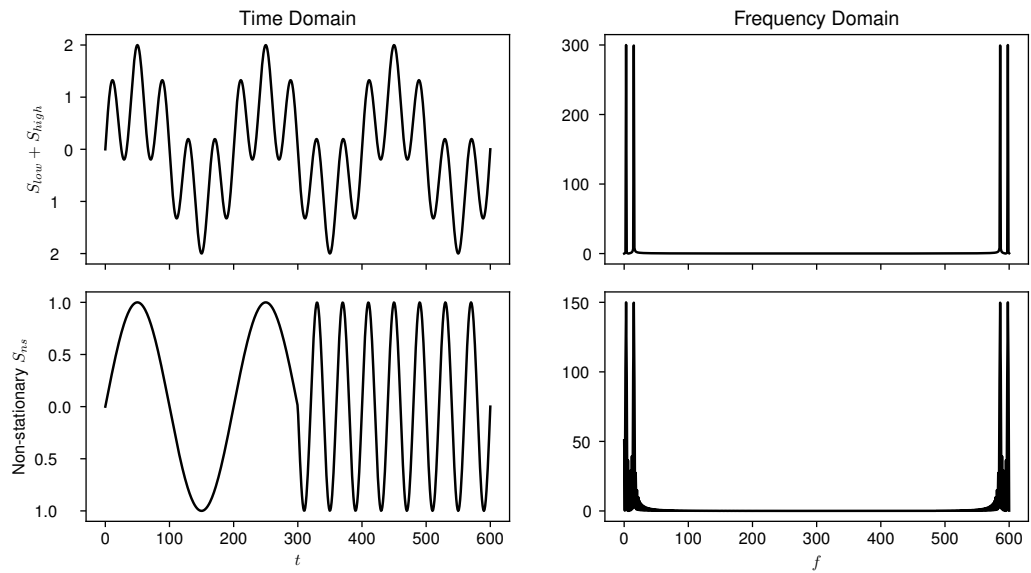}}
  \caption{The changes over time/space of a signal are not captured by the FFT.}
  \label{fig:spatial_resolution_fft}
\end{figure}

\begin{figure}[H]
  \begin{subfigure}[t]{.33\textwidth}
    \centering
    \includegraphics{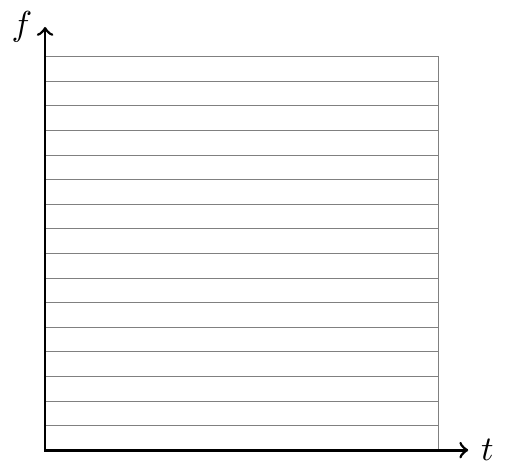}
    \captionsetup{width=0.8\textwidth}
    \caption{Fourier Transform, time resolution is the duration of the signal.}
    \label{fig:grid_a_fft}
  \end{subfigure}
  \begin{subfigure}[t]{.33\textwidth}
    \centering
    \includegraphics{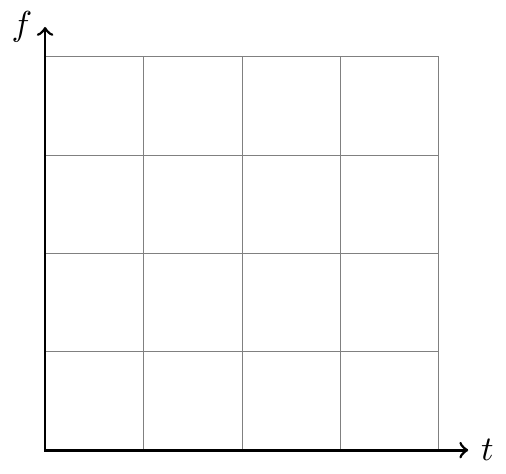}
    \captionsetup{width=0.8\textwidth}
    \caption{Windowed-FFT or Short-Time-FFT, fixed width window}
    \label{fig:grid_b_windowed}
  \end{subfigure}
  \begin{subfigure}[t]{.33\textwidth}
    \centering
    \includegraphics{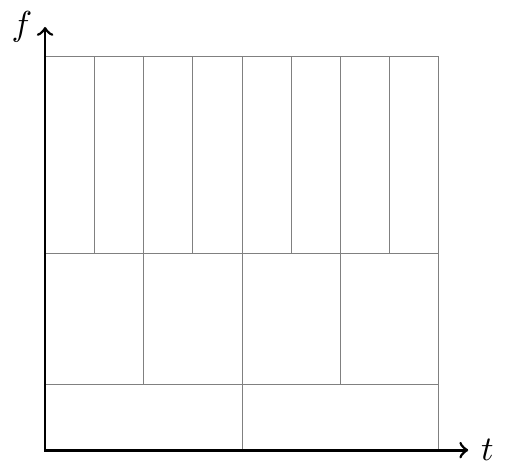}
    \captionsetup{width=0.8\textwidth}
    \caption{Discrete Wavelet Transform. Adaptive window, better time resolution at high frequencies.}
    \label{fig:grid_c_wavelet}
  \end{subfigure}
  \caption{Simultaneous spatial and frequency resolution of different transformations. $\Delta t, \Delta f$}
  \label{fig:grids}
\end{figure}

A few examples to illustrate the concept: In a discrete case, such as an image, this means that the FFT has the highest resolution on frequency, but where, in the spatial domain, a specific frequency occurs is reduced to `somewhere', bounded by the size of the image (\autoref{fig:spatial_resolution_fft}, \ref{fig:grid_a_fft}).
Or in signals occurring only in one time-lapse: they can't be differentiated from the same signal occurring continuously, so they will share similar spectra representation (\autoref{fig:spatial_resolution_fft}). In the same line, non-stationary signals which frequency depends on time will give a spectrum where all those frequencies are represented, but there is no information about when, in time, the change of frequency occurred.

The short-time FFT add spatial/time localization dividing the original signal in small, consecutive segments, and applying the FFT to each of those. The problem with this approach, besides being computationally expensive, is that events shorter than the time window are still not resolved and that the width of the segment is constant (\autoref{fig:grid_b_windowed}). Would it be possible to have better spatial resolution for some components of the frequency spectra, such as high frequency components (\autoref{fig:grid_c_wavelet})?

\subsection{Wavelet transformation}
\label{sub:transform}

Let consider the following wavelet decomposition $\forall f \in L_2(\mathbb{R}^d)$
\begin{equation}
    f(\bm{x}) = \sum_{s \in \mathbb{Z}} \sum_{\bm{l} \in \mathbb{Z}^d} \langle f,\psi_{s,\bm{l}} \rangle \conj\psi_{s,\bm{l}}(\bm{x})
\end{equation}
The family of functions $\{\psi_{s,\bm{l}}\}$ is a wavelet \textit{frame} (Def. \ref{def:frame}), constructed by means of \textit{translations} and \textit{dilations} (Def. \ref{def:D_T}) of the \textit{mother wavelet} function $\psi$.
\begin{equation}
    \psi_{s,\bm{l}}(\bm{x}) = \abs{\det(A)}^{-s/2}\psi(A^s\bm{x} - \bm{l})
\label{eq:wavelet-frame}
\end{equation}
Each dilation, defined by the dilation matrix $A$, squeezes or stretches the mother wavelet, acting as a change of scale. The translation operator moves and centers the location of the mother wavelet $\psi$. If the dilation matrix is diagonal with the same dilation factor $a$ in all dimensions, $A = a \mathcal{I}_d$, then \autoref{eq:wavelet-frame} becomes:
\begin{equation}
    \psi_{s,\bm{l}}(\bm{x}) = a^{-d\cdot s/2}\psi(a^s\bm{x} - \bm{l})
\end{equation}

The mother wavelet function has to have finite energy $\psi \in L_2(\mathbb{R}^d)$, i.e $\int_{-\infinity}^{\infinity}\abs{\psi(\bm{x})}^2 d\bm{x} < \infinity $.\newline

\begin{definition}[Dilations and Translations]\label{def:D_T}
  Given a function $f\in\mathbb{R}$ we define the dilation and translation operators\cite{heil_continuous_1989}:\par
  \begin{tabular}{lll}
    Dilation:&  $D_a f(x):=\abs{a}^{-1/2}f(x/a)$ &for $a \in \mathbb{R}\backslash\{0\}$ \\
    Translation:&   $T_b f(x):=f(x-b)$ &for $b \in \mathbb{R}$ \\
  \end{tabular}

  $\mathbb{R}^d$ generalization:\\
  Given $f \in L_2(\mathbb{R}^d), f:\mathbb{R}^d \rightarrow \mathbb{R}$, $\bm{x}\in\mathbb{R}^d$, the dilation scalar $a$ is replaced by a dilation matrix $A = a \mathcal{I}_d$ \cite{qian_wavelet_2007}. The dilation matrix $A$ is expansive, having all its eigenvalues $\abs{\lambda_i} > 1$, so it is invertible. For the translation operator, the scalar $b$ is replaced for the vector $\bm{b}$.\par
  \begin{tabular}{lll}
    Dilation:& $D_A f(\bm{x}):=\abs{\det(A)}^{-1/2}f(A\bm{x})$ &$A \text{ expansive matrix}$\\
    Translation:&   $T_{\bm{b}} f(\bm{x}):=f(\bm{x}-\bm{b})$ &for $\bm{b} \in \mathbb{R}^d$ \\
  \end{tabular}
\end{definition}

\begin{definition}[Frame]\label{def:frame}
    A family of functions $\{\phi_{\bm{k}}\}_{ \bm{k}\in\mathbb{Z}^d}$ is a \textit{frame} of $L_2(\mathbb{R}^d)$ if and only if there exists two positive constants A, B $< \infinity$ such that:

\begin{equation}
  A\norm{f}^2 \le \sum_{\bm{k}\in\mathbb{Z}^d}\abs{\langle\phi_{\bm{k}},f\rangle}^2 \le B\norm{f}^2, \forall f \in L_2(\mathbb{R}^d)
\end{equation}

The frame is \textit{tight} if $A=B$. If $A=B=1$ we have a \textit{Parseval frame} that satisfies the decomposition/reconstruction formula:
\begin{equation}
  f = \sum_{\bm{k}\in\mathbb{Z}^d}\langle\phi_{\bm{k}},f\rangle \cdot \phi_{\bm{k}}, \forall f \in L_2(\mathbb{R}^d)
\end{equation}
which has the same form than the expansion using an orthonormal basis, however in the frame generalization, the family $\phi_{\bm{k}}$ may be redundant.
\end{definition}

\subsection{Wavelet Pyramid}
\label{sub:wavelet_pyramid}

\begin{figure}[H]
  \begin{subfigure}[t]{\textwidth}
    \centering
    \includegraphics{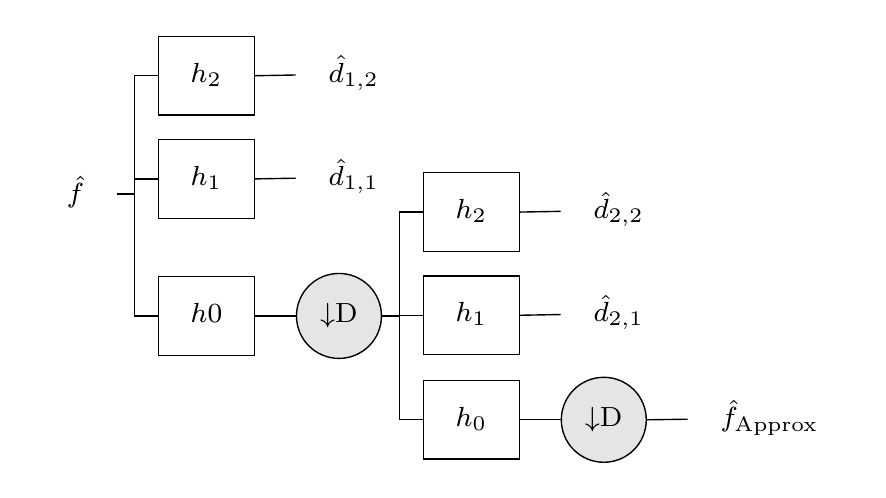}
    \captionsetup{width=0.8\textwidth}
    \caption{forward/analysis.}
    \label{fig:pyramid_analysis_a}
  \end{subfigure}
  \vspace*{\floatsep}
  \begin{subfigure}[t]{\textwidth}
    \centering
    \includegraphics{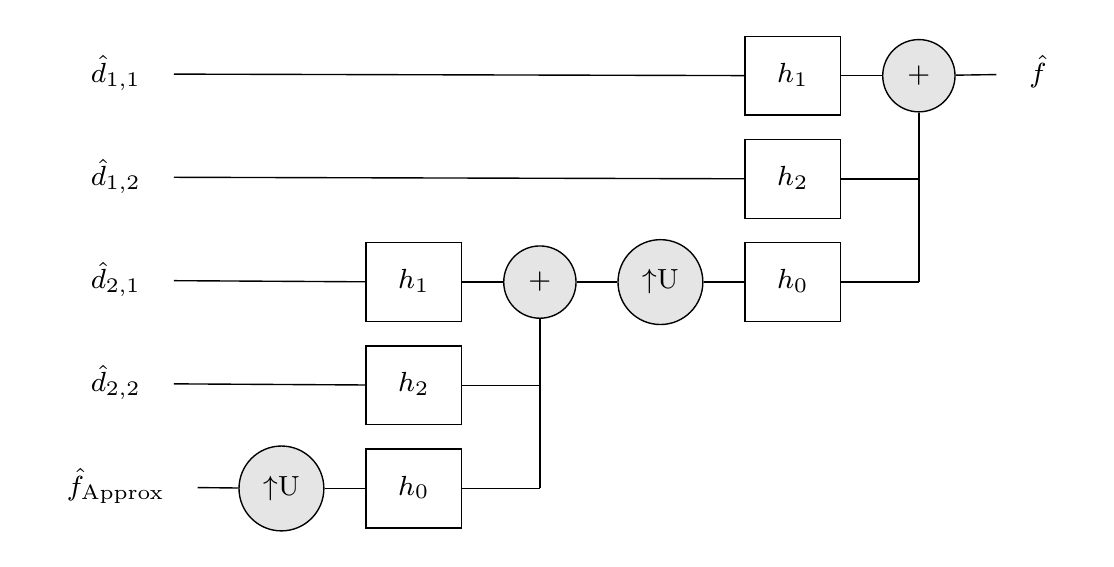}
    \captionsetup{width=0.8\textwidth}
    \caption{inverse/reconstruction.}
    \label{fig:pyramid_reconstruction_b}
  \end{subfigure}
  \caption{Forward \ref{fig:pyramid_analysis_a} and Inverse wavelet \ref{fig:pyramid_reconstruction_b} two-level pyramid with two high pass sub-bands.}
  \label{fig:wavelet_pyramid}
\end{figure}

A band-limited pyramid is created by applying, at each level, a low-pass filter $h_0$ and downsampling by a scale factor of two, see \autoref{fig:pyramid_analysis_a}. Because the wavelet frames are tight, we can get perfect reconstruction applying the inverse pyramid from \autoref{fig:pyramid_reconstruction_b}. In \autoref{fig:wavelet_pyramid} we use two levels and two high pass subbands $h_1, h_2$. The results of this pyramid are the detail coefficients $d_{s,h}$, where $s \in \{1,\ldots, \text{Levels}\}$, and $h \in \{1,\ldots, \text{HighPassSubBands}\}$. Note that these details are in the frequency domain, if the original image was given in spatial domain, an inverse Fourier transform for each $d_{s,h}$ must be performed to get the wavelet coefficients in the spatial domain.

\begin{figure}[H]
  \centering
  \includegraphics{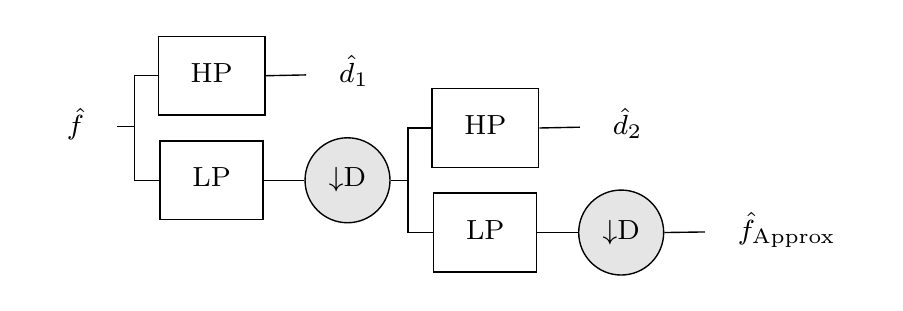}
  \caption{Forward wavelet pyramid with a classic two level filter bank and only one sub-band, HP is the high pass filter, and LP low pass}
  \label{fig:pyramid_simple_2}
\end{figure}

\begin{figure}[H]
  \begin{subfigure}[t]{.333\textwidth}
    \centering
    \includegraphics[width=\textwidth]{images/phaseImages/held2D_6_1.png}
    \captionsetup{width=\textwidth}
    \caption{Level: 6, HighPassSubBands: 1}
    \label{fig:phase_bands61}
  \end{subfigure}
  \begin{subfigure}[t]{.333\textwidth}
    \centering
    \includegraphics[width=\textwidth]{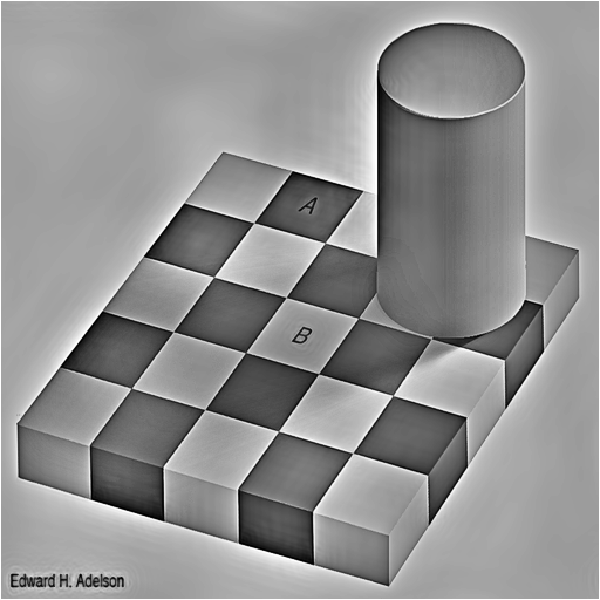}
    \captionsetup{width=\textwidth}
    \caption{HighPassSubBands: 2}
    \label{fig:phase_bands62}
  \end{subfigure}
  \begin{subfigure}[t]{.333\textwidth}
    \centering
    \includegraphics[width=\textwidth]{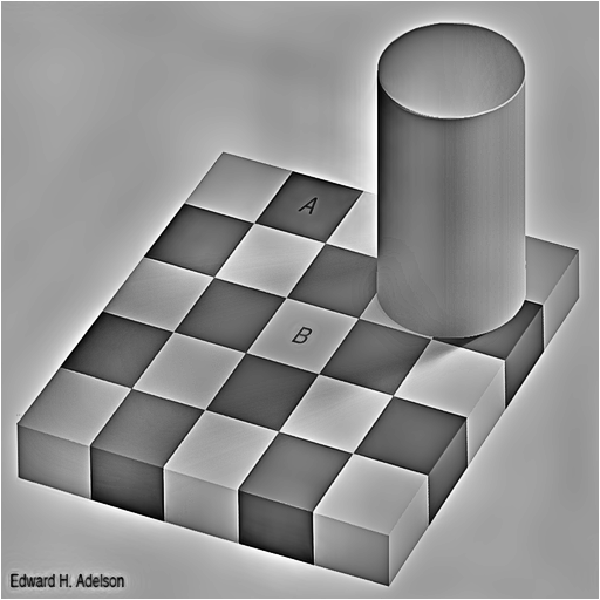}
    \captionsetup{width=\textwidth}
    \caption{HighPassSubBands: 3}
    \label{fig:phase_bands63}
  \end{subfigure}
  \vspace*{\floatsep}
  \begin{subfigure}[t]{.333\textwidth}
    \centering
    \includegraphics[width=\textwidth]{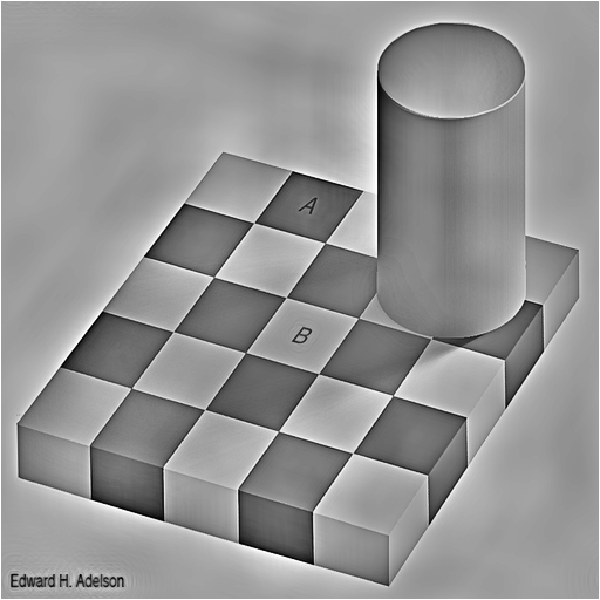}
    \captionsetup{width=\textwidth}
    \caption{HighPassSubBands: 4}
    \label{fig:phase_bands64}
  \end{subfigure}
  \begin{subfigure}[t]{.333\textwidth}
    \centering
    \includegraphics[width=\textwidth]{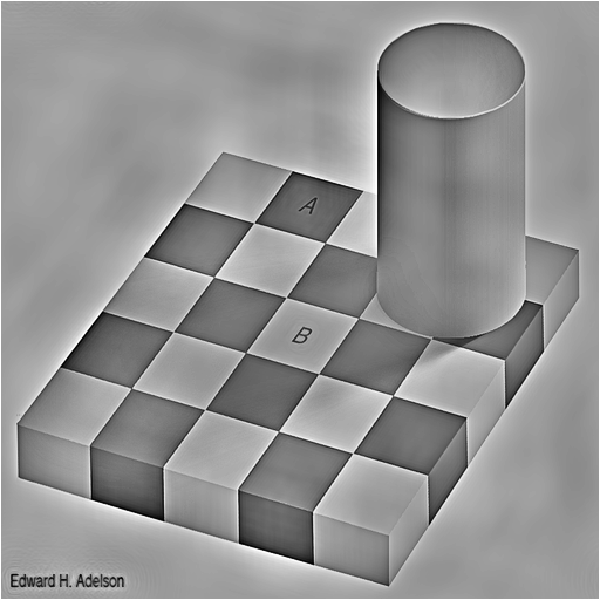}
    \captionsetup{width=\textwidth}
    \caption{HighPassSubBands: 5}
    \label{fig:phase_bands65}
  \end{subfigure}
  \begin{subfigure}[t]{.333\textwidth}
    \centering
    \includegraphics[width=\textwidth]{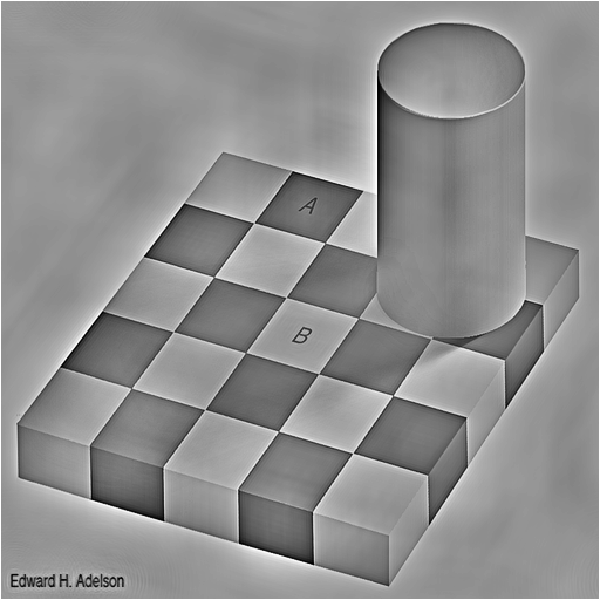}
    \captionsetup{width=\textwidth}
    \caption{HighPassSubBands: 10}
    \label{fig:phase_bands610}
  \end{subfigure}
  \caption{Using six scales (Level: 6), results for different number of high frequency sub-bands.}
  \label{fig:phase_bands}
\end{figure}

Usually in the literature use only one high-pass sub-band, the wavelet filter bank consists then in one high pass filter and one low pass filter (see\autoref{fig:pyramid_simple_2}). The adventages of the subbands is that they provide more frequency resolution \cite{held_steerable_2010}, at expenses of more computation time, and also they might generate spatial domain distortions due to multiple sharp cutoffs. For phase detection (see the application in \autoref{fig:phase_bands}), they are fundamental.

\subsection{Isotropic wavelets}
\label{sub:isotropic_wavelets}
These wavelets are non-separable in Cartesian coordinates, depending on $\norm{\bm{\omega}}$. Isotropic wavelets have the same mother wavelet for each scale. All of the implemented wavelets fulfill the conditions of Prop \ref{prop:tight_conditions}, their shape differ in the spatial decay, and vanishing moments. Which one is better depend on the application, and developing new wavelets is a research topic. We have implemented four, there are at least \cite{unser_steerable_2011} two isotropic wavelets missing: Papadakis \cite{romero_geometry_2009}, and Meyer \cite{daubechies_ten_1992}.
The advantage of isotropic wavelets is that they are steerable, but in order to get directionality features, the wavelets are coupled with other filters that gather directional information, such as the Riesz transform, see \autoref{sec:riesz_transform}.

\begin{figure}[H]
  \centering
  \begin{subfigure}[t]{.49\textwidth}
    \centering
    \includegraphics[width=0.99\textwidth]{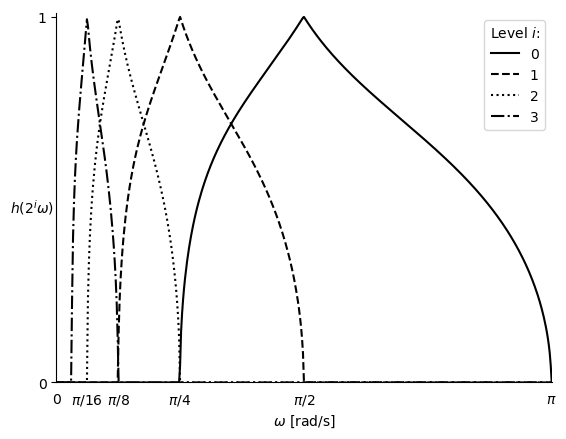}
    \captionsetup{width=\textwidth}
    \caption{Vow.}
    \label{fig:mother_vow}
  \end{subfigure}
  \begin{subfigure}[t]{.49\textwidth}
    \centering
    \includegraphics[width=0.99\textwidth]{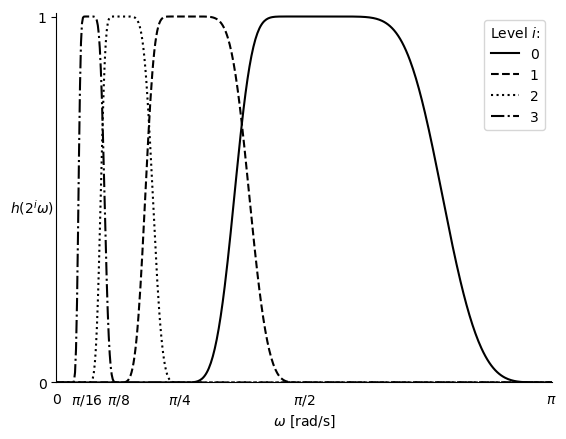}
    \captionsetup{width=\textwidth}
    \caption{Held.}
    \label{fig:mother_held}
  \end{subfigure}
  \vspace*{\floatsep}
  \begin{subfigure}[t]{.49\textwidth}
    \centering
    \includegraphics[width=0.99\textwidth]{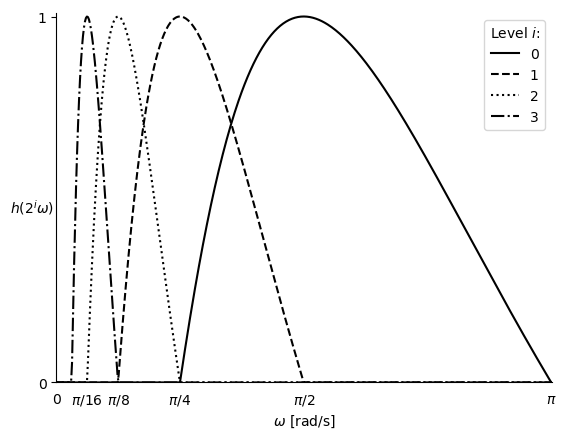}
    \captionsetup{width=\textwidth}
    \caption{Simoncelli.}
    \label{fig:mother_simoncelli}
  \end{subfigure}
  \begin{subfigure}[t]{.49\textwidth}
    \centering
    \includegraphics[width=0.99\textwidth]{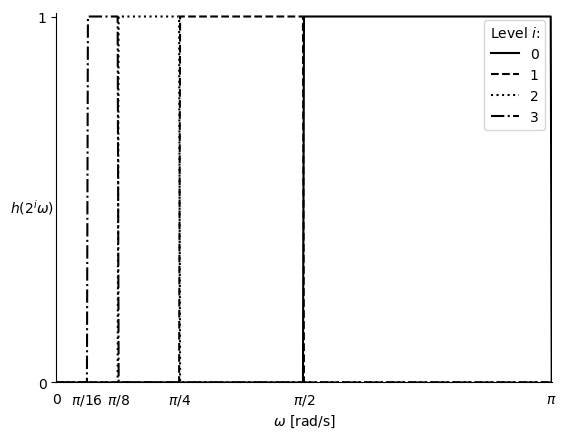}
    \captionsetup{width=\textwidth}
    \caption{Shannon.}
    \label{fig:mother_shannon}
  \end{subfigure}
  \caption{Tiling of the frequency domain by isotropic wavelets when the dilation factor is $2$. All the wavelets fulfill the conditions from Proposition \ref{prop:tight_conditions}. The mother wavelets are represented at $i=0$. }
  \label{fig:mother_wavelets}
\end{figure}

The present requirement that the mother wavelet is isotropic constraints the wavelet-design. The goal is to generate a tight wavelet frame and have perfect reconstruction with the inverse pyramid.
\begin{proposition}\label{prop:tight_conditions}
  Conditions for the mother wavelet $\psi$ to generate a tight wavelet frame \cite{held_steerable_2010, unser_steerable_2011, chenouard_3d_2012}:\newline
Let $h(w)$ be a radial frequency profile such that:
\begin{description}[font=\normalfont , leftmargin=*, labelsep=*, labelindent=*]
    \item[1. (Band-limitedness):] $h(\w) = 0, \forall \w > \pi$.
    \item[2. (Riesz Partition of Unity):] $\sum_{i \in \mathbb{Z}} \abs{h\Bigl(\Bigl(\bigl(\trans{A}\bigr)^{-1}\Bigr)^i\w\Bigr)}^2 = 1$.\newline
          If $A = a \mathcal{I}_d$: $\sum_{i \in \mathbb{Z}} \abs{h(a^i\w)}^2 = 1$. \cite{held_steerable_2010, aldroubi_wavelets_2007}
     \item[3. (Vanishing Moments):] $\dv[n]{h(\w)}{\w}\Bigr|_{\w=0} = 0, \text{for } n = 0, \ldots, N$.
\end{description}
If the mother wavelet $\psi$ is defined by $\fft{\psi}(\bm{\w})= h(\norm{\bm{\w}})$, then it generates a tight wavelet frame in $L_2(\mathbb{R}^d)$, where $\fft{\psi}$ is the ND Fourier transform of $\psi$.

Condition 2 guarantees that the tiling from all the scales fills the frequency domain, see \autoref{fig:mother_wavelets}.
\end{proposition}

\noindent\begin{minipage}[t]{0.49\textwidth}
  \textbf{VOW}, variance-optimal wavelets \cite{pad_vow:_2014} :
\begin{equation*}
\label{VOW}
  h(\omega) =
    \begin{cases}
    \begin{aligned}
      &\sqrt{\frac{1}{2} + \frac{\tan(\kappa(1+2\log_2\frac{2\omega}{\pi}))}{2\tan(\kappa)}} , &\omega \in [\frac{\pi}{4} , \frac{\pi}{2} [ \\
      &\sqrt{\frac{1}{2} - \frac{\tan(\kappa(1+2\log_2\frac{\omega}{\pi}))}{2\tan(\kappa)}} , &\omega \in [\frac{\pi}{2} , \pi ] \\
      &0, &\text{otherwise}
    \end{aligned}
    \end{cases}
\end{equation*}
\begin{equation*}
  \text{where } \kappa \in [0, \frac{\pi}{2}] \text{ is found to be } 0.75
\end{equation*}
\newline
\end{minipage}
\noindent\begin{minipage}[t]{0.49\textwidth}
  \textbf{Held} \cite{held_steerable_2010} :
\begin{equation*}
\label{Held}
  h(\omega) =
    \begin{cases}
    \begin{aligned}
&\cos\left(2\pi q_n(\frac{\omega}{2\pi})\right) , &\omega \in ]\frac{\pi}{4} , \frac{\pi}{2} ] \\
  &\sin\left(2\pi q_n(\frac{\omega}{4\pi})\right) , &\omega \in ]\frac{\pi}{2} , \pi ] \\
      &0, &\text{otherwise}
    \end{aligned}
    \end{cases}
\end{equation*}
\begin{equation*}
  \text{where } q_n \text{ is a polynomial function of order n } 
\end{equation*}
\end{minipage}
\noindent\begin{minipage}[t]{0.49\textwidth}
  \textbf{Simoncelli} \cite{portilla_image_2000, simoncelli_steerable_1995} :
\begin{equation*}
\label{Simoncelli}
  h(\omega) =
    \begin{cases}
    \begin{aligned}
  &\cos\left(\frac{\pi}{2} \log_2\frac{2\omega}{\pi}\right) , &\omega \in ]\frac{\pi}{4} , \frac{\pi}{2} ] \\
  &0, &\text{otherwise}
    \end{aligned}
    \end{cases}
\end{equation*}
\end{minipage}
\noindent\begin{minipage}[t]{0.49\textwidth}
\textbf{Shannon:}
\begin{equation*}
\label{Shannon}
  h(\omega) =
    \begin{cases}
    \begin{aligned}
  &1, &\omega \in [\frac{\pi}{2} , \pi ] \\
  &0, &\text{otherwise}
    \end{aligned}
    \end{cases}
\end{equation*}
\end{minipage}

\begin{figure}[H]
  \centering
  \begin{subfigure}[t]{.49\textwidth}
    \centering
    \includegraphics[width=0.99\textwidth]{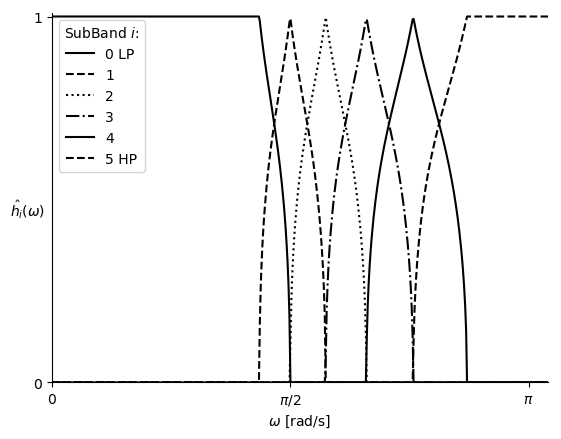}
    \captionsetup{width=\textwidth}
    \caption{Vow.}
    \label{fig:subbands_vow}
  \end{subfigure}
  \begin{subfigure}[t]{.49\textwidth}
    \centering
    \includegraphics[width=0.99\textwidth]{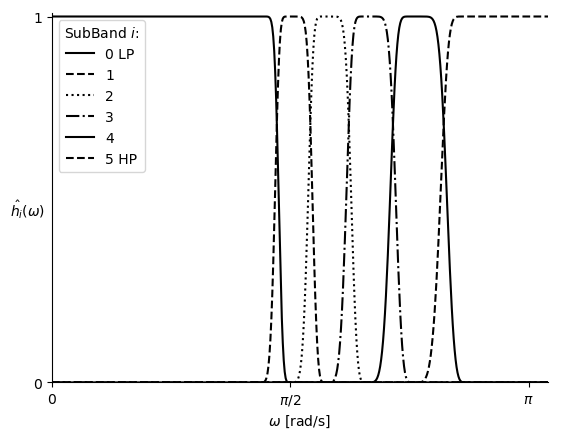}
    \captionsetup{width=\textwidth}
    \caption{Held.}
    \label{fig:subbands_held}
  \end{subfigure}
  \vspace*{\floatsep}
  \begin{subfigure}[t]{.49\textwidth}
    \centering
    \includegraphics[width=0.99\textwidth]{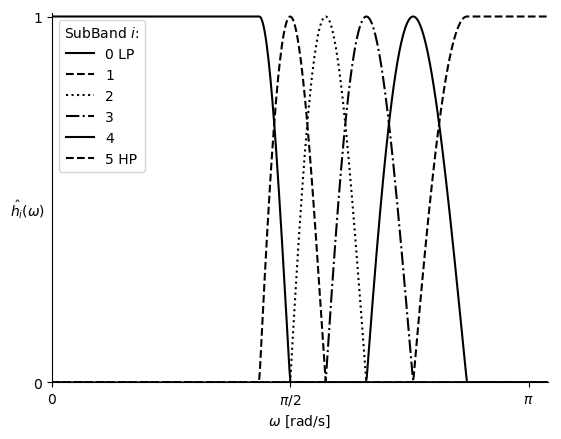}
    \captionsetup{width=\textwidth}
    \caption{Simoncelli.}
    \label{fig:subbands_simoncelli}
  \end{subfigure}
  \begin{subfigure}[t]{.49\textwidth}
    \centering
    \includegraphics[width=0.99\textwidth]{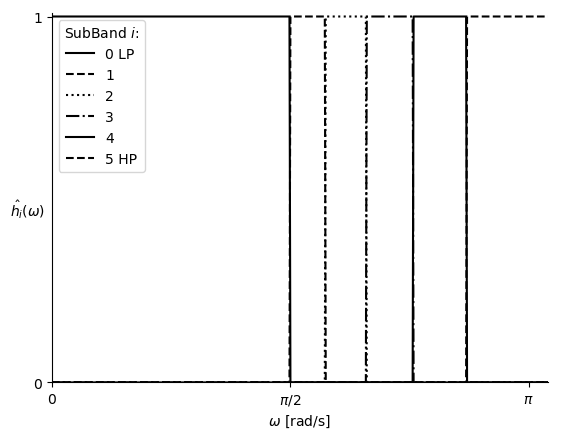}
    \captionsetup{width=\textwidth}
    \caption{Shannon.}
    \label{fig:subbands_shannon}
  \end{subfigure}
  \caption{Shape of SubBands when HighPassSubBands = 5, sub-bands increase the frequency resolution, but can generate extra artifacts in the spatial domain due to the sharp frequency cut-offs.}
  \label{fig:subbands_wavelets}
\end{figure}

\section{Riesz Transform}
\label{sec:riesz_transform}
The forward or analysis wavelet pyramid outputs a set of wavelet coefficients with information about each scale.
A steerable filter \cite{held_steerable_2010, unser_steerable_2011, simoncelli_steerable_1995} can be applied to the output of the wavelet pyramid. This is the main purpose of the isotropy of the mother wavelet, the steerable filter is used to select the orientation where the feature of interest is maximum.

There are mathematical constraints in the steerable filters that can be coupled with the wavelet pyramid. Read more: \cite{unser_multiresolution_2009, unser_steerable_2011}.

One of the transform that can be coupled to the pyramid is a Riesz transform $\bm{\mathcal{R}}$ of order $N=1$
\begin{equation}
\bm{\mathcal{R}} f(\bm{x}) =
  \begin{pmatrix}
    \mathcal{R}_{1} f(\bm{x}) \\
    \vdots \\
    \mathcal{R}_d f(\bm{x})
  \end{pmatrix}
  \hspace{0.6cm}
  \xleftrightarrow[\mathcal{F}^{-1}]{\hspace{0.4cm}\mathcal{F}\hspace{0.4cm}}
  \hspace{0.6cm}
  \bm{\hat{\mathcal{R}}}\hat{f}(\bm{\omega}) = -j \frac{\bm{\omega}}{\norm{\bm{\omega}}}\hat{f}(\bm{\omega})
\end{equation}
where $d$ is the image dimension and the number of components of the Riesz transform of order $N=1$. $j=\sqrt{-1}$ and $\mathcal{F}, \mathcal{F}^{-1}$ are the forward and inverse Fourier Transform.\newline

The Riesz transform is the generalization of the Hilbert transform for ND. The Hilbert transform $\mathcal{H}$ is used to generate the \textit{analytic signal} in 1D.
\begin{equation}
\label{eq:analytic-signal}
\begin{aligned}
  f_a(x) &= f(x) + j \mathcal{H}f(x) \\
  \mathcal{H}f(x) = (h * f)(x)
  \hspace{0.6cm}
  &\xleftrightarrow[\mathcal{F}^{-1}]{\hspace{0.4cm}\mathcal{F}\hspace{0.4cm}}
  \hspace{0.6cm}
  \hat{\mathcal{H}}\hat{f}(\omega) = -jsgn(\omega)\hat{f}(\omega) = -j \frac{\omega}{\abs{\omega}}\hat{f}(\omega)
\end{aligned}
\end{equation}

\subsection{Monogenic Signal}
\label{sub:monogenic}

There is no direct generalization of the analytical signal in ND, but we can generate a signal with similar properties, called the \textbf{Monogenic signal} using the Riesz transform of order 1\cite{felsberg_monogenic_2001, kovesi_image_1999}.
The Monogenic signal is a set formed by the original signal and the D-components of the first order Riesz transform.
\begin{equation}
  f_a = \{f, \mathcal{R}_x, ..., \mathcal{R}_d\}
\end{equation}

The amplitude $A$ and the phase $P$ at each location of the Monogenic signal are:
\begin{equation}
\label{eq:monogenic_amplitude}
 A(\mathbf{x_0})= \sqrt{f(\mathbf{x_0})^2 + A_R(\mathbf{x_0})^2 }
\end{equation}
where $ A_R(\mathbf{x_0})=  \sqrt{\sum_{i=1}^N \mathcal{R}_i(\mathbf{x_0})^2} $
\begin{equation}
\label{eq:monogenic_phase}
P(\mathbf{x_0})= \text{atan2}(A_R(\mathbf{x_0}),A(\mathbf{x_0}))
\end{equation}

The Monogenic signal can be used to perform local phase analysis for feature detection in ND.

\subsection{Generalized Riesz Transform}
\label{sub:generalized_riesz}
The Riesz transform will map any frame of $L_2(\mathbb{R}^d)$ into another one \cite{held_steerable_2010, unser_wavelet_2010}. This mapping property allows to couple the Riesz transform of any order with the wavelet pyramid and if the wavelet function is isotropic, to get perfect reconstruction when performing the inverse transform. The Riesz transform of order $N$ generate a vector containing smoothed directional Nth derivatives.

We will summarize the multiindex notation introduced in \cite{unser_steerable_2011}.\newline
Consider $ \vn = (n_{1}, \ldots, n_{d})$ a d-dimensional multiindex vector, where the $n_i$ entries are non-negative integers. And then define the following operators and operations:
\begin{enumerate}[topsep=0pt]
  \item Sum of components: $\abs{\bm{n}} = \sum_{i=1}^d n_i = N$.
  \item Factorial: $\vn! = n_1!n_2!\cdots n_d!$
  \item Exponentiation of a vector $z = (z_i,\cdots,z_d) \in \mathbb{C}^d$:
    $z^{\vn} = z_{1}^{n_1} \cdots z_{d}^{n_d}$
  \item Partial derivative of a function $f(\vx), \vx = (x_1,\ldots, x_d) \in \mathbb{R}^d$:
    $\partial^{\vn}f(\vx) = \frac{\partial^{N}f(\vx)}{\partial x_{1}^{n_1}\cdots \partial x_{d}^{n_d}}$
\end{enumerate}

Riesz transforms can be connected to the partial derivatives \cite{unser_steerable_2011}:
\begin{equation*}
\label{eq:riesz_partial_derivatives}
\mathcal{R}^{\vn}(-\Delta)^{\frac{\abs{\vn}}{2}}f(\vx) = (-1)^{\abs{\vn}}\sqrt{\frac{\abs{\vn}!}{\vn!}}\partial^{\vn}f(\vx)
\end{equation*}
where $(-\Delta)^\gamma$ is the fractional Laplace of order $\gamma$. From \cite{unser_steerable_2011}: \blockquote{Since the inverse of $(-\Delta)^{\frac{\abs{\vn}}{2}}$ is an isotropic low-pass-filtering operator, the net effect of the higher order Riesz transform is to extract smoothed version of the derivatives of order N of the signal of interested.}

The number of components of $\bm{\mathcal{R}}$ depends on the order of the Riesz transform $N$, and the dimension of the signal $d$:
\begin{equation}
\label{eq:riesz_components}
 M = p(N,d) = \frac{(N+d-1)!}{(d-1)! N!}
\end{equation}

\subsection{Generalized Steerable Framework}
Simoncelli \cite{simoncelli_steerable_1995} coupled the multi-resolution analysis using wavelets with the steerable concept developed years before \cite{freeman_design_1991}. A steerable framework can be used with polar separable functions. In this case, the radial part of the function has to be calculated only once, and the polar part has only to be computed in those directions that constitute a basis of the space. After this, the filter can be oriented, or steered to any direction, using a weighted linear combination of the basis. This approach gives a directional analysis but keeping it computationally performant.

The steerable framework can be combined with wavelets when these are isotropic, which implies polar separability.
Simoncelli's framework has been used in a lot of 2D applications. Freeman in its seminal work \cite{freeman_design_1991} wrote a generalized version for any dimension, but it was lately, Chenouard and Unser \cite{held_steerable_2010,unser_steerable_2011,chenouard_3d_2012} who used rotation matrices in 3D.

\begin{definition}
A function $f$ is steerable in three dimensions if it can be expressed as:
\begin{equation}
  f(\bm{Rx}) = \sum_{m=1}^M k_m(\bm{R})g_m(\bm{x})
\end{equation}
where $\bm{x} = (x_1,x_2,x_3)$ is any 3D vector in Cartesian coordinates and $\bm{R}$ is any rotation matrix in three dimensions.
$\{g_m\}_{m=1...M}$ is the set of primary functions, ie , a basis, and $\{k_m\}_{m=1...M}$ is a set of interpolation functions with $M < \infinity$.
\end{definition}

\section{Implementation Details}
\label{sec:Impl}
\subsection{Summary}
\label{sub:Summary}

Most of the filters in this module require input images in the dual space (frequency domain). If working with regular spatial-domain images, a \doxygen{ForwardFFTImageFilter} has to be applied. The decision is based on performance and accuracy, avoiding expensive convolution operations and also multiple Fourier transforms.

Also, because we work in the frequency domain, we add a \github{FrequencyShrinkImageFilter} and a \github{FrequencyExpandImageFilter} \textbf{without} any interpolation. The shrinker chops the high frequency pixels of the image. And the expander adds zeros in the higher frequency bins.
These filters require input images to be hermitian --note that the output of a forward FFT on a real image is hermitian.\newline
General FrequencyExpanders and FrequencyShrinkers that can be applied to non-hermitian complex images are not implemented, although some preliminary work can be found in \url{https://github.com/phcerdan/ITKIsotropicWavelets/pull/31} based on \cite{portilla_image_2003}.

\subsection{Frequency Iterators}
\label{sub:frequencyiterators}
Every filter that uses the frequency value of a pixel can be templated with a \github{FrequencyImageRegionIterator}.\newline
This kind of iterators add the functions GetFrequencyBin(), GetFrequencyIndex() to a regular \doxygen{ImageRegionIterator}, helping to abstract the complexity of the frequency layout into the iterator.
The layout determines the order and location of the frequency bins: zero frequency --DC component--, Nyquist, low/high, or positive and negative frequencies. The layout changes depending on the parity of the image, the forward Fourier transform algorithm chosen, or if the frequencies have been shifted (\doxygen{FFTShiftImageFilter}).\newline
Also these frequency iterators have data members: FrequencyOrigin, FrequencySpacing, holding metadata information about the frequency domain.\newline
ITK puts a strong emphasis in the ImageInformation or metadata on spatial domain images: Origin, Spacing, Direction, Index. There are strong requirements when dealing with images with different metadata, for example, multiplication between two images is only possible when they have compatible ImageInformation.\newline
Although, when applying a \doxygen{ForwardFFTImageFilter}, the output image, which is now in the frequency domain, still have metadata referring to the spatial domain, but nothing about the frequency origin, or spacing between frequencies.

There are three classes of FrequencyImageRegionIterator:
\begin{description}[topsep=0pt,font=\normalfont\textbullet\space]
  \item [FFTLayout:] \github{FrequencyFFTLayoutImageRegionIterator}\newline
    This iterator assumes that the frequency image has a ``standard layout''. Different libraries, such as VNL, FFTW, numpy.fft, generate this layout. Each image dimension is divided in two regions, holding positive and negative frequencies.
    \begin{description}[topsep=0pt,font=\normalfont-\space]
      \item[ZeroFrequency:] The index holding the zero frequency value is at the origin $[0,\ldots,0]$ (upper-left, corner).
      \item[Nyquist:] When size $N$ is even, the positive Nyquist frequency is located in the middle, index $N/2$, and the negative Nyquist is not stored.
    When $N$ is odd, there is no Nyquist frequency, but most positive frequency is at index $(N-1)/2$, and most negative frequency is at index $(N+1)/2$
    \end{description}
    The first index after the origin corresponds to the positive lowest frequency, say $0.1 \text{Hz}$. If this is the index $1$, the latest index $N-1$ corresponds to the less negative frequency $-0.1 \text{Hz}$

  \item [ShiftedFFTLayout:] \github{FrequencyShiftedFFTLayoutImageRegionIterator}\newline
    The standard layout can be confusing to reason about, a common alternative is to shift the zero frequency bin to the center of the image via \doxygen{FFTShiftImageFilter}.
  \item [Regular:] \github{FrequencyImageRegionIterator}.\newline
    This iterator is for images that were taken experimentally in the frequency domain. GetFrequency() here is just a wrap of TransformIndexToPhysicalPoint, and GetFrequencyBin is equal to GetIndex. It assumes that the image metadata refers to the frequency domain, so FrequencyOrigin and FrequencySpacing are equal to the regular Origin and Spacing.
    Iterators with GetFrequency() functions are needed for other classes in the module.
\end{description}

This abstraction will hopefully facilitate the implementation of further filters manipulating images in the frequency domain.

\subsection{Wavelet Transform}
\label{sub:wavelet_transform_impl}

See \autoref{sub:wavelet_pyramid}. The Wavelet Transform is templated over one of the IsotropicWavelet functions from \autoref{sub:isotropic_wavelets} chosen by the user through a \github{WaveletFrequencyFilterBankGenerator}. The generator uses the function to generate an image or a filter bank.

We can compute the maximum levels that an input accept with the ComputaMaxNumberOfLevels, right now the implementation accepts needs inputs of the form $s^M$, where $s$ is the scale factor chosen for peforming the multiresolution, and M is an integer. Even though this condition is restrictive, we can resize any input using \doxygen{FFTPadImageFilter}, or if the user pipeline involves neighbor iterators, prefer \github{FFTPadPositiveIndexImageFilter}, implemented in this module, that avoids setting negative indices.

The \github{WaveletFrequencyForward} generates a set of coefficients generated after applying the high pass filters $h_{\text{band}}$ for each level, and an approximation (the result of applying the cascade of low pass filter $h_0$).

The wavelet coefficients from the forward pyramid can be manipulated to perform further image analysis, for example edge detection, denoise, phase analysis for feature detection, etc. These extra analysis should be independent of the multiresolution framework.

After any manipulation, we can perform an inverse pyramid to reconstruct the image. Doing this require to plug in the modified wavelet coefficients into \github{WaveletFrequencyInverse}, if the wavelet coefficients are not modified by any further analysis, we get exactly the same image as the origin image that we input to the forward pyramid.

\subsection{Riesz Transform}
\label{sub:riesz_transform_impl}

We have a \github{RieszFrequencyFunction}, implementing a Generalized Riesz Function \cite{unser_steerable_2011} of any order $N>0$. This function receives an input $\in \mathcal{R}^d$ and outputs a vector of M Riesz Components, see \autoref{eq:riesz_components}. To generate images from this function, we use the \github{RieszFrequencyFilterBankGenerator}. To rotate or steer the result of the Riesz transform,use \github{RieszRotationMatrix}. Please be aware the multiindex notation in these classes, see \autoref{sub:generalized_riesz}

\subsection{Structure Tensor}
\label{sub:structure_tensor_impl}

Given an array of inputs, \github{StructureTensor} \cite{unser_steerable_2011} computes the linear combination (or direction) of inputs that maximizes the response for each location in the image. Instead of only measuring the response at the pixel of interest, it takes into account a local neighborhood. \cite{unser_steerable_2011}.

$$
 \mathbf{u}({\mathbf{x}_0}) = \arg \max_{\Vert\mathbf{u}\Vert =1 }  \int_{\mathbb{R}^d} g(\mathbf{x} - \mathbf{x}_0) \left| \mathbf{I}_{\mathbf{u}}(\mathbf{x})\right|^2
$$
$$
 \left| \mathbf{I}_{\mathbf{u}}(\mathbf{x})\right|^2 =
 \mathbf{u}^T \cdot \mathbf{I}(\mathbf{x}) \cdot (\mathbf{I}(\mathbf{x}))^T \cdot \mathbf{u}
$$
 $ \mathbf{I}$ is the required std::vector of input images. These images might be the output of a directional filter to an image
 (for example, directional derivatives from an image) or the basis of a steerable filter, such as a RieszImageFilter.
 Instead of just selecting the max response from the vector at every pixel, it uses the response over
 a local neighborhood, specified using an isotropic Gaussian window $g(\mathbf{x})$.
 This approach is more robust against noise. The user can control the radius and sigma of this Gaussian kernel.
 Estimation of the local orientation this way results in an eigen-system with matrix:
 $$
 [\mathbf{J}(\mathbf{x}_0)]_{mn} = \sum_{\mathbf{x} \in \mathbb{Z}^d} g(\mathbf{x} - \mathbf{x}_0) I_m[\mathbf{x}]I_n[\mathbf{x}]
 $$
 where $I_m, I_n $ are input images, $ m,n \in {0,N-1} $ and $N$ is the total number of inputs.
 $g$ is a Gaussian kernel.

 The solution of the EigenSystem defined by $\mathbf{J}$ are the N EigenValues and EigenVectors.
 The output of StructureTensor is a 2D Matrix of size (N,N+1), where the submatrix (N,N) are the EigenVectors, and the last column (N+1) are the EigenValues.
 The orientation that maximizes the response: $\bm{u}$ is the EigenVector with largest EigenValue, which is is the Nth column of the output matrix.
 We can use the calculated direction $\bm{u}$ to get a new image with max response from the inputs at each pixel with the function ComputeProjectionImageWithLargestResponse(),
 or any other direction from other eigen vectors with ComputeProjectionImage

 Also we can compare eigen values to study the local coherency of each pixel:
 $$
 \chi (\mathbf{x}_0)= \frac{\lambda_N(\mathbf{x}_0) - A(\mathbf{x}_0)}{\lambda_N(\mathbf{x}_0) + A(\mathbf{x}_0)}
 $$
where $\lambda_N(\mathbf{x}_0)$ is the largest eigen value at pixel $\mathbf{x}_0$ ,
and
$A(\mathbf{x}_0) = \frac{1}{N-1}\sum_{i=1}^{N-1}\lambda_i(\mathbf{x}_0)$ is the average of the other eigen values.

\subsection{Phase Analysis}
\label{sub:phase_analyzers}

This module also implements a base class \github{PhaseAnalysisImageFilter}, and a specialization \github{PhaseAnalysisSoftThresholdImageFilter} that applies a soft-threshold technique to ignore low amplitude values. The example application \ref{fig:phase_hermann}, uses a \hyperref[sub:monogenic]{Monogenic signal} to study the local phase of each wavelet coefficient to perform multi-scale feature detection \cite{felsberg_monogenic_2001,unser_multiresolution_2009}.

\section{A guided example:}
I recommend the reader interested in extend this module to have a look to the tests for more usage options.
\subsection{Input in the frequency domain.}
The input for the \github{WaveletFrequencyForward} has to be a complex image of a float/double pixel type. If the image is not in the frequency domain already, apply a \doxygen{ForwardFFTImageFilter} to get it.

Some FFT algorithms only work for specific image sizes, for example, the size has to be a power of two.
You can apply a \doxygen{FFTPadImageFilter} to pad the image with zeros to reach the required size. However, this filter sets some padded areas with negative indices, which can generate problems with neighbor iterators.
As a work around, this module introduces \github{FFTPadPositiveIndexImageFilter} to avoid negative indices.

\subsection{Choosing an isotropic wavelet}
\label{sub:Choosing}
Choose a mother wavelet from the options available \ref{sub:isotropic_wavelets}, and create a
\github{WaveletFrequencyFilterBankGenerator} with the type of the mother wavelet as a template parameter.

\begin{minted}{c++}
  typedef itk::HeldIsotropicWavelet< PixelType, ImageDimension>
    WaveletFunctionType;
  typedef itk::WaveletFrequencyFilterBankGenerator< ComplexImageType, WaveletFunctionType >
    WaveletFilterBankType;
\end{minted}

\subsection{Forward / Analysis}
\label{sub:Forward}
Perform the wavelet transform based on input levels and high-frequency sub bands.
\inline{ComputeMaxNumberOfLevels} is a static class function to calculate the max level.
Currently, the size of input image has to be a multiple of two, but the implementation has the member \inline {m_ScaleFactor} for future extension and relaxation of this constraint. Use FFTPadPositiveIndexImageFilter for padding the image with zeros for a valid size.

\begin{minted}{c++}
  typedef itk::WaveletFrequencyForward< ComplexImageType, ComplexImageType, WaveletFilterBankType >
    ForwardWaveletType;
  typename ForwardWaveletType::Pointer forwardWavelet = ForwardWaveletType::New();
  forwardWavelet->SetHighPassSubBands( highSubBands );
  forwardWavelet->SetLevels(levels);
  forwardWavelet->SetInput(fftFilter->GetOutput());
\end{minted}

\subsection{Inverse / Reconstruction}
\label{sub:Inverse}
Before applying the inverse wavelet and get a reconstructed image, you want to modify the wavelet coefficients first. The design of the framework has tried to decouple both pyramids, so we can focus in algorithms that deal only the wavelet coefficients, on not with the details of the multiresolution framework.\par
Right now, only a phase analysis filter has been developed, but there are plenty of room for more, see \autoref{sec:Conclusion}.
Set the same options used in the forward pyramid. And set the modified wavelet coefficients. If you don't modify them you will reconstruct the original image in the frequency domain.

Perform the wavelet forward with a suitable image:
\begin{minted}{c++}
#include "itkHeldIsotropicWavelet.h"
#include "itkVowIsotropicWavelet.h"
#include "itkSimoncelliIsotropicWavelet.h"
#include "itkShannonIsotropicWavelet.h"
#include "itkWaveletFrequencyFilterBankGenerator.h"
#include "itkWaveletFrequencyForward.h"
...

  const unsigned int ImageDimension = 3;
  typedef double                                          PixelType;
  typedef std::complex< PixelType >                       ComplexPixelType;
  typedef itk::Image< ComplexPixelType, ImageDimension >  ComplexImageType;

  // Set the WaveletFunctionType and the WaveletFilterBank
  typedef itk::HeldIsotropicWavelet< PixelType, ImageDimension >
        WaveletFunctionType;
  typedef itk::WaveletFrequencyFilterBankGenerator< ComplexImageType, WaveletFunctionType >
    WaveletFilterBankType;

  // WaveletFrequencyForward
  typedef itk::WaveletFrequencyForward< ComplexImageType, ComplexImageType, WaveletFilterBankType >
    ForwardWaveletType;
  typename ForwardWaveletType::Pointer forwardWavelet = ForwardWaveletType::New();
  forwardWavelet->SetHighPassSubBands( highSubBands );
  forwardWavelet->SetLevels(levels);
  forwardWavelet->SetInput(fftFilter->GetOutput());
  forwardWavelet->Update();

  // Result of the wavelet decomposition.
  typename ForwardWaveletType::OutputsType analysisWaveletCoeffs =
    forwardWavelet->GetOutputs();
\end{minted}

With the wavelet coefficients in hand (in the frequency domain), perform a decoupled filter, phase analysis, denoising, feature detection, etc:
\begin{minted}{c++}
  // Manipulate coefficients for your purposes:
  // Denoise, phase analysis, etc...
  typename ForwardWaveletType::OutputsType modifiedWaveletCoeffs;
  for( unsigned int i = 0; i < forwardWavelet->GetNumberOfOutputs(); ++i )
    {
    ...
    aWaveletCoefficientFilter->SetInput( analysisWaveletCoeffs[i] );
    aWaveletCoefficientFilter->Update();
    modifiedWaveletCoeffs.push_back( aWaveletCoefficientFilter->GetOutput() );
    }
\end{minted}

And then plug back the modified coefficients to the inverse pyramid to get a reconstruction.
\begin{minted}{c++}
  // Inverse Wavelet Transform
  typedef itk::WaveletFrequencyInverse< ComplexImageType, ComplexImageType, WaveletFilterBankType >
    InverseWaveletType;
  typename InverseWaveletType::Pointer inverseWavelet = InverseWaveletType::New();
  inverseWavelet->SetHighPassSubBands( highSubBands );
  inverseWavelet->SetLevels( levels );
  // inverseWavelet->SetInputs( forwardWavelet->GetOutputs() );
  inverseWavelet->SetInputs( modifiedWaveletCoeffs );
  bool useWaveletFilterBankPyramid = true;
  inverseWavelet->SetUseWaveletFilterBankPyramid( useWaveletFilterBankPyramid );
  inverseWavelet->SetWaveletFilterBankPyramid( forwardWavelet->GetWaveletFilterBankPyramid() );
  inverseWavelet->Update();

  // The output of the inverse wavelet is in the frequency domain.
  // Perform an InverseFFT to get a real image.
  typename InverseFFTFilterType::Pointer inverseFFT = InverseFFTFilterType::New();
  inverseFFT->SetInput( inverseWavelet->GetOutput() );

  // Write or visualize the reconstructed output.
  writer->SetInput(inverseFFT->GetOutput());

\end{minted}

\subsection{PhaseAnalyzer}

Here we show the results of the test \href{https://github.com/phcerdan/ITKIsotropicWavelets/blob/5c9f8db0718164675eecd6b134e66d4015394ff8/test/itkRieszWaveletPhaseAnalysisTest.cxx}{itkRieszWaveletPhaseAnalysisTest} with a couple of optical illusions images, a checker board and a Hermann grid.

\begin{figure}[H]
  \begin{subfigure}[t]{.49\textwidth}
    \centering
    \includegraphics[width=\textwidth]{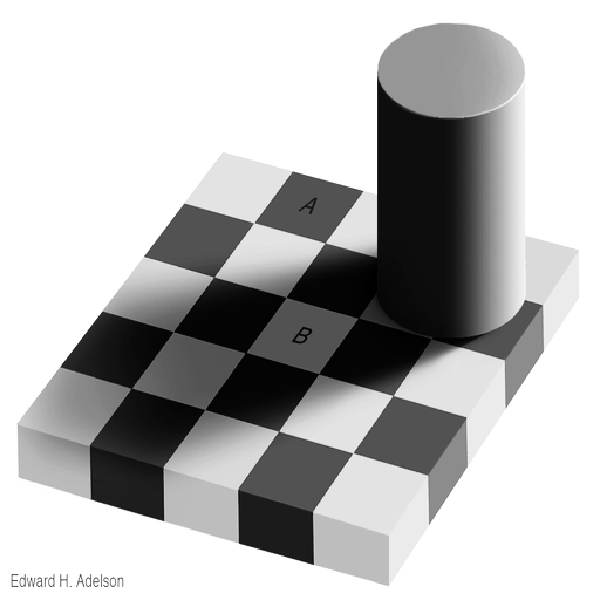}
    \captionsetup{width=\textwidth}
    \caption{Original}
    \label{fig:phase_original}
  \end{subfigure}
  \begin{subfigure}[t]{.49\textwidth}
    \centering
    \includegraphics[width=\textwidth]{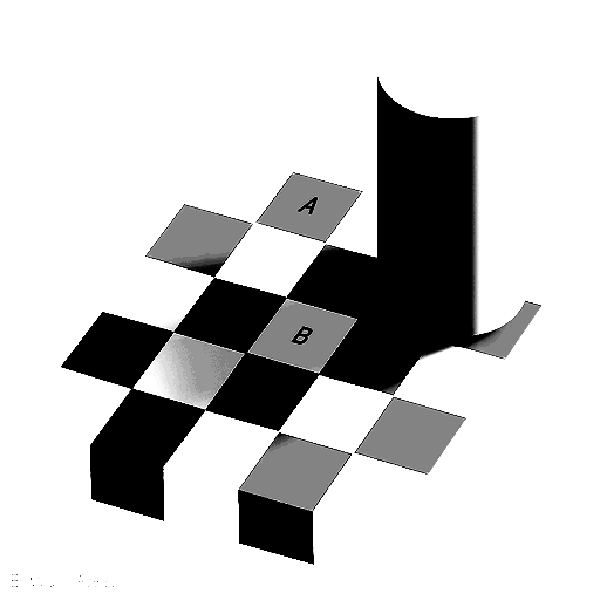}
    \captionsetup{width=\textwidth}
    \caption{Intensity thresholding}
    \label{fig:phase_grayscale}
  \end{subfigure}
  \caption{The original image looks like a regular check-board, but it isn't. Pixels in the regions A and B have the same intensity value (129), however our vision system performs local phase analysis that allows us to treat A,B regions as different, keeping a global checker-board structure. \ref{fig:phase_grayscale} uses a non-linear map of intensity-color to enhance the irregular checker board structure.}
  \label{fig:phase_checker}
\end{figure}

\begin{figure}[H]
  \begin{subfigure}[t]{.333\textwidth}
    \centering
    \includegraphics[width=\textwidth]{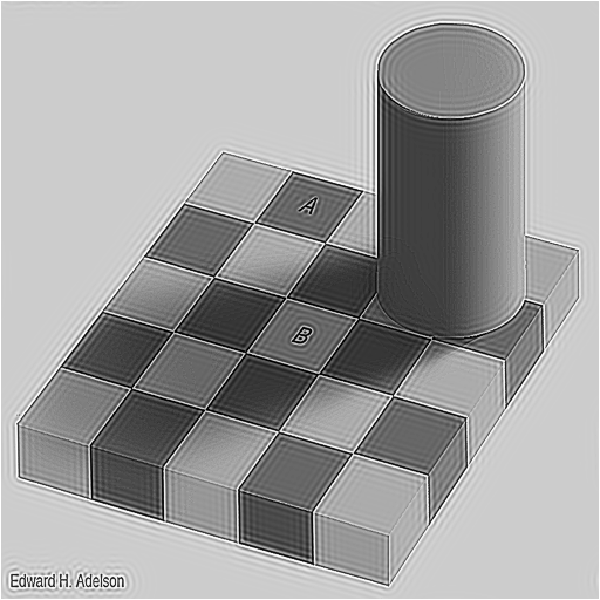}
    \captionsetup{width=\textwidth}
    \caption{Levels: 1}
    \label{fig:phase11}
  \end{subfigure}
  \begin{subfigure}[t]{.333\textwidth}
    \centering
    \includegraphics[width=\textwidth]{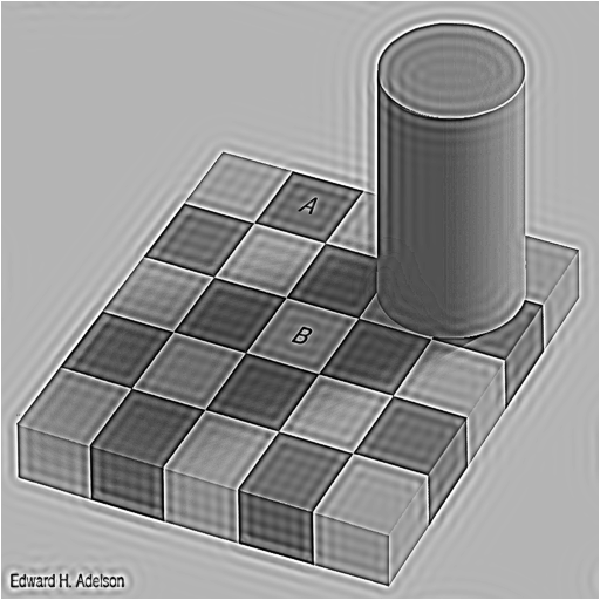}
    \captionsetup{width=\textwidth}
    \caption{Levels: 2}
    \label{fig:phase21}
  \end{subfigure}
  \begin{subfigure}[t]{.333\textwidth}
    \centering
    \includegraphics[width=\textwidth]{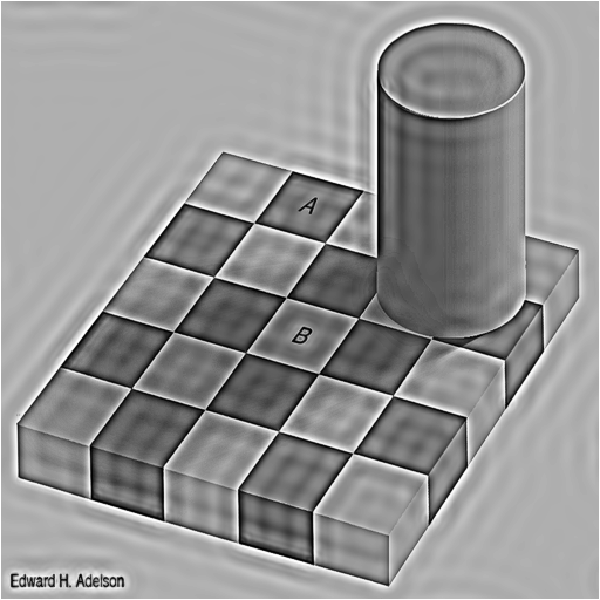}
    \captionsetup{width=\textwidth}
    \caption{Levels: 3}
    \label{fig:phase31}
  \end{subfigure}
  \vspace*{\floatsep}
  \begin{subfigure}[t]{.333\textwidth}
    \centering
    \includegraphics[width=\textwidth]{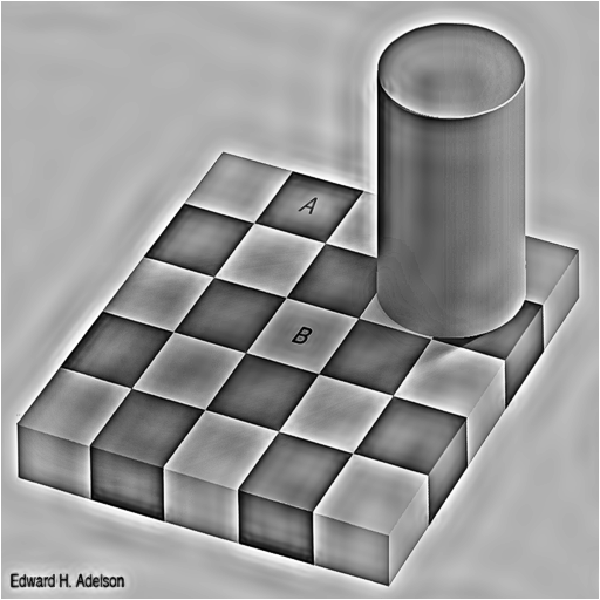}
    \captionsetup{width=\textwidth}
    \caption{Levels: 4}
    \label{fig:phase41}
  \end{subfigure}
  \begin{subfigure}[t]{.333\textwidth}
    \centering
    \includegraphics[width=\textwidth]{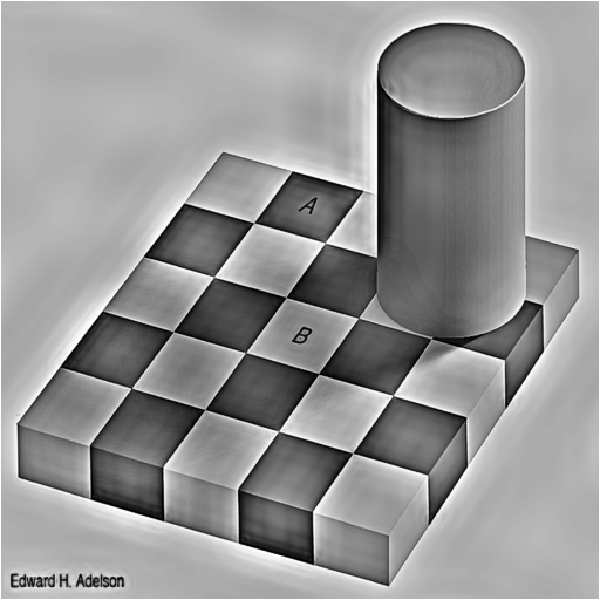}
    \captionsetup{width=\textwidth}
    \caption{Levels: 5}
    \label{fig:phase51}
  \end{subfigure}
  \begin{subfigure}[t]{.333\textwidth}
    \centering
    \includegraphics[width=\textwidth]{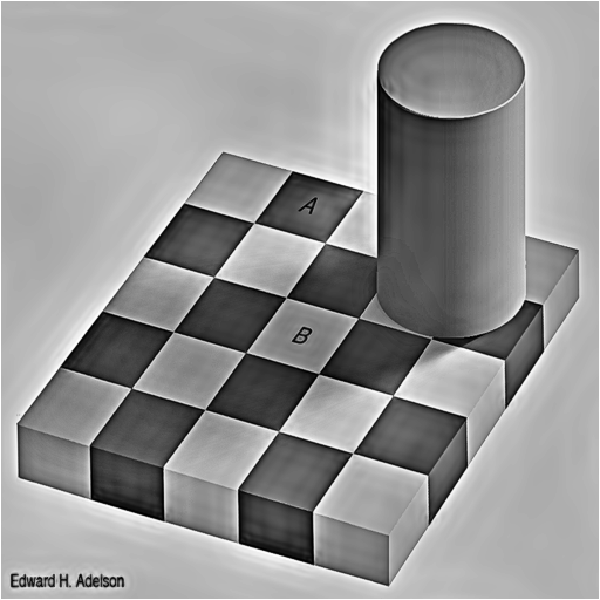}
    \captionsetup{width=\textwidth}
    \caption{Levels: 6 }
    \label{fig:phase61}
  \end{subfigure}
  \caption{(Results of the phase analysis (with soft threshold) for different number of scales in the wavelet pyramid. The input image is a checker-board of size $512x512$). }
  \label{fig:phase_levels}
\end{figure}

\begin{figure}[H]
  \centering
  \begin{subfigure}[t]{0.4\textwidth}
    \centering
    \includegraphics[width=0.8\textwidth]{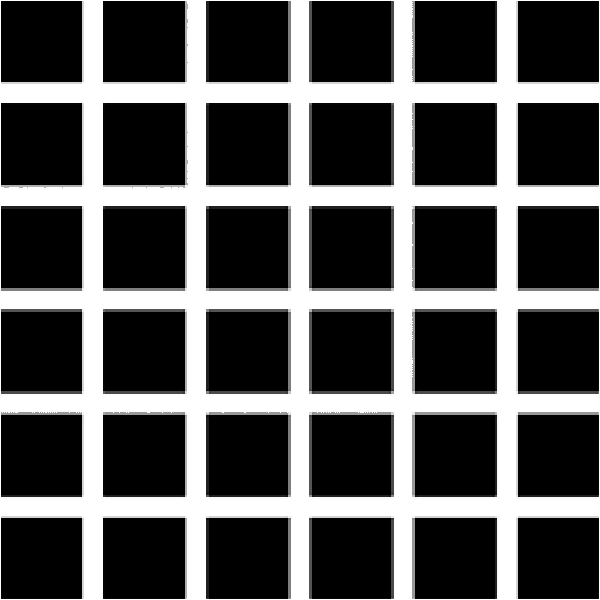}
    \captionsetup{width=\textwidth}
    \caption{Hermann Grid}
    \label{fig:hermann_original}
  \end{subfigure}
  \vspace*{\floatsep}
  \begin{subfigure}[t]{.4\textwidth}
    \centering
    \includegraphics[width=0.8\textwidth]{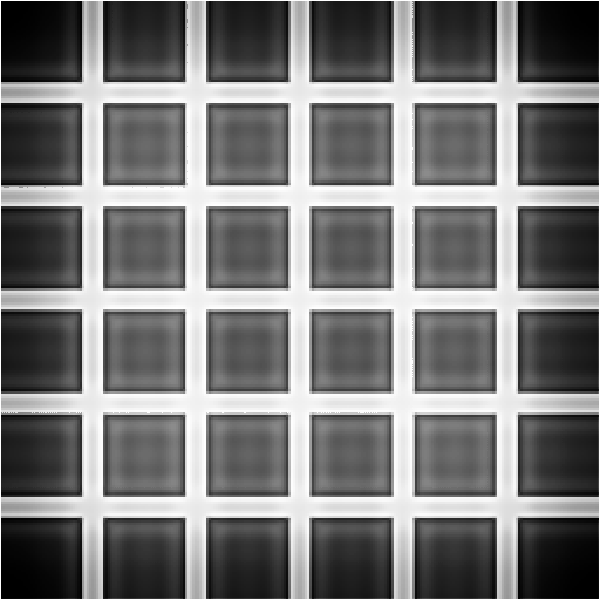}
    \captionsetup{width=\textwidth}
    \caption{Levels: 8 (max), HighPassSubBands: 1}
    \label{fig:hermann_8_1}
  \end{subfigure}
  \begin{subfigure}[t]{.4\textwidth}
    \centering
    \includegraphics[width=0.8\textwidth]{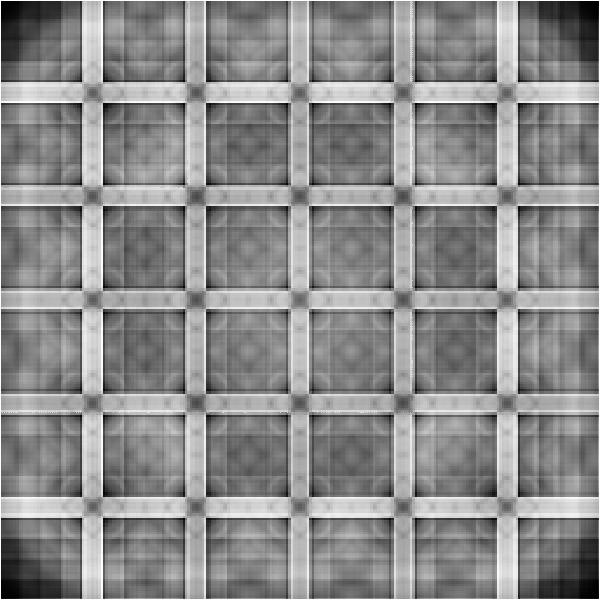}
    \captionsetup{width=\textwidth}
    \caption{Levels: 8 (max), HighPassSubBands: 10}
    \label{fig:hermann_8_10}
  \end{subfigure}
  \caption{The optical illusion generated by \ref{fig:hermann_original} the Hermann grid is generated by the local phase analysis of our vision. \ref{fig:hermann_8_1} is the result of a phase analysis with the Monogenic signal for a wavelet pyramid of 8 levels and only one high pass sub band. \ref{fig:hermann_8_10} is the same pyramid but with 10 high pass bands to increase the frequency resolution. Both results use Simoncelli wavelet, but there is not much difference using Held, or Vow mother wavelets.}
  \label{fig:phase_hermann}
\end{figure}
\section{Conclusion and future work}
\label{sec:Conclusion}

This work provides ITK with a multiresolution analysis based on wavelet decomposition using isotropic and steerable wavelets. Also it provides utilities to work in the dual or frequency domain. Frequency iterators, shrinkers and expanders. It also provides a sub-sampler without interpolation, and a expander with zeros that work in any domain.

As an application, we showed in \autoref{fig:phase_hermann} a local phase analyzer using the wavelet coefficients based on \cite{held_steerable_2010}.

Future work would be easier to implement with the tools already developed. Some work will only require plumbing operations together in a new filter, for example, creating a specific class for the Simoncelli Steerable Framework \cite{simoncelli_steerable_1995}, and the more general Steerable framework using Riesz transform \cite{unser_steerable_2011}, that will use the already implemented \github{RieszRotationMatrix}.

More applications that will require extra work but use the same wavelet backbone are:
\begin{itemize}[topsep=0pt]
  \item Denoising algorithms can be developed using the exposed wavelet coefficients, for example: Gaussian Scale Mixture Models (GSM)\cite{portilla_image_2003}, or SURE-LET \cite{blu_sure-let_2007}.
  \item Feature detection without using template matching in an efficient way \cite{puspoki_template-free_2015}.
\end{itemize}

 Implementation of these algorithms will provide state of the art solutions based on wavelets to a wider audience, and I am happy to link them from the External Repository, \url{https://github.com/phcerdan/ITKIsotropicWavelets} to have a reference of wavelet solutions for ITK.

 Already implemented but not used in any application yet are: \github{RieszRotationMatrix}, providing a steerable framework for the General Riesz Transform \cite{unser_steerable_2011}. And \github{StructureTensor}, a local estimator of the direction --steer-- with the highest contribution \cite{chenouard_3d_2012}.

\section{Acknowledgements}
\label{sec:Acknow}
I would like to thank Jon Haitz Legarreta for improving tests and documentation of the module.

%

%
%
\twocolumn
\bibliography{Wavelet}
\onecolumn

\end{document}